\documentclass{article}

\usepackage[utf8]{inputenc} 
\usepackage[T1]{fontenc}    
\usepackage{booktabs}       
\usepackage{amsfonts}       
\usepackage{nicefrac}       
\usepackage{microtype}      
\usepackage{xcolor}         
\usepackage{times}  
\usepackage{helvet}  
\usepackage{courier}  
\usepackage[hyphens]{url}  
\usepackage{graphicx} 
\usepackage[numbers]{natbib}  
\newcommand{\citetnew}[1]{\citeauthor{#1} (\citeyear{#1})}
\usepackage{caption} 
\usepackage[preprint]{neurips_2024}

\usepackage{algorithm}%
\usepackage{algorithmicx}%
\usepackage{algpseudocode}%
\usepackage{amsmath}
\usepackage{amsfonts}
\usepackage{booktabs}
\usepackage{multirow}
\usepackage{tabularx}
\usepackage{makecell}
\usepackage{pifont}
\usepackage{newfloat}
\usepackage{listings}
\usepackage{adjustbox}

\usepackage{wrapfig}
\usepackage{amssymb}
\usepackage{caption}
\usepackage[table]{xcolor}

\DeclareCaptionStyle{ruled}{labelfont=normalfont,labelsep=colon,strut=off} 
\floatstyle{ruled}
\newfloat{listing}{tb}{lst}{}
\floatname{listing}{Listing}

\title{GRAD: Graph-Retrieved Adaptive Decoding for Hallucination Mitigation}
\author{%
  Manh~Nguyen, 
  Sunil~Gupta, 
  Dai~Do,
  Hung~Le\\
  Applied Artificial Intelligence Initiative \\
  Deakin University\\
  Geelong, Australia \\
  \texttt{\{manh.nguyen, sunil.gupta, v.do, thai.le\}@deakin.edu.au} \\
}

\begin{document}

\maketitle

\begin{abstract}

Hallucination mitigation remains a persistent challenge for large language models (LLMs), even as model scales grow. Existing approaches often rely on external knowledge sources, such as structured databases or knowledge graphs, accessed through prompting or retrieval. However, prompt-based grounding is fragile and domain-sensitive, while symbolic knowledge integration incurs heavy retrieval and formatting costs. Motivation from knowledge graph, we introduce \textbf{Graph-Retrieved Adaptive Decoding (GRAD)}, a decoding-time method that grounds generation in corpus-derived evidence without retraining. GRAD constructs a sparse \textbf{token transition graph} by accumulating next-token logits across a small retrieved corpus in a single forward pass. During decoding, graph-retrieved logits are max-normalized and adaptively fused with model logits to favor high-evidence continuations while preserving fluency. Across three models and a range of question-answering benchmarks spanning intrinsic, extrinsic hallucination, and factuality tasks, GRAD consistently surpasses baselines, achieving up to 9.7\% higher intrinsic accuracy, 8.6\% lower hallucination rates, and 6.9\% greater correctness compared to greedy decoding, while attaining the highest truth–informativeness product score among all methods. GRAD offers a lightweight, plug-and-play alternative to contrastive decoding and knowledge graph augmentation, demonstrating that statistical evidence from corpus-level token transitions can effectively steer generation toward more truthful and verifiable outputs.

\end{abstract}

\section{Introduction}

Large language models (LLMs) demonstrate impressive reasoning and compositional abilities but remain prone to \textit{hallucinations}, fluent yet factually incorrect statements \citep{ji2023survey}. This issue is particularly severe in open-domain or retrieval-augmented generation, where models must synthesise information from noisy, incomplete, or low-precision retrieved contexts \cite{niu2024ragtruth}. As LLMs are increasingly deployed in knowledge-critical applications, mitigating hallucinations has become an urgent research goal.

Existing approaches can be broadly categorized into three groups. First, training-time alignment methods, such as reinforcement learning from human feedback \citep{ouyang2022training} or AI feedback \citep{bai2022constitutional}, fine-tune models toward truthful or safe responses. While effective, they are computationally costly and tied to specific datasets. Second, prompt-based or in-context learning (ICL) methods guide models toward factual reasoning via prompt engineering or self-verification \citep{dai2022promptagator, manakul2023selfcheckgpt}. However, they depend on fragile instructions, are sensitive to example order and quality \citep{nori2023can}, and often overfit to particular domains. Third, decoding-time interventions adjust token generation dynamically without retraining. Contrastive or entropy-based methods, such as CAD \citep{shi2024trusting}, DoLa \citep{chuang2024dola}, and Instructive Decoding \citep{instructivedecoding} modify next-token logits by suppressing untruthful or uncertain signals. Although efficient, these methods often rely on heuristic contrasts or internal activations that fail to generalize to long or noisy contexts.

Parallel efforts attempt to ground generation using external knowledge graphs (KGs) or verifiable databases \citep{baek2023knowledge, sun2023think, wen2024mindmap}. While explicit and interpretable, these symbolic structures require repeated LLM–KG interactions for retrieval, schema alignment, and reasoning, leading to high computational overhead and latency. Moreover, prompt-based KG integration increases input length and sensitivity to prompt format, reducing scalability and robustness.

We propose \textbf{Graph-Retrieved Adaptive Decoding (GRAD)}, a novel decoding framework that grounds generation in \textit{statistical co-occurrence evidence} from a retrieved corpus. Unlike symbolic KGs, GRAD builds a \textbf{token transition graph} $\mathcal{G}$ by accumulating raw model logits over question-answer pairs in a single forward pass. Each directed edge $\mathcal{E}(u,v)$ captures the cumulative predictive confidence that token $v$ follows token $u$, forming a corpus-level prior over likely continuations. During decoding, GRAD retrieves the corresponding graph logits, normalizes them, and adaptively fuses them with model logits, biasing generation toward high-evidence continuations while maintaining flexibility in uncertain regions.

Recent studies such as \citetnew{bang-etal-2025-hallulens} emphasize that most hallucination research remains constrained to factuality-oriented QA benchmarks like TruthfulQA, which risk overfitting to a single dataset and conflate factuality with broader hallucination types. To address this, new benchmarks have been introduced to systematically evaluate both intrinsic (internal consistency) and extrinsic (knowledge-grounded) hallucinations. We adopt this broader evaluation perspective to examine GRAD's ability to mitigate diverse hallucination forms. We evaluate GRAD across three challenging benchmarks: \textbf{FaithEval} (intrinsic consistency under noisy contexts), \textbf{PreciseWikiQA} (extrinsic factuality with graded difficulty), and \textbf{WikiQA} (truth-informativeness trade-off), using \textbf{Qwen2.5 (1.5B/3B)} and \textbf{Llama3.2–3B}. GRAD consistently outperforms greedy decoding and four strong baselines, improving intrinsic accuracy by up to 9.7\%, reducing hallucination rates by up to 8.6\%, and enhancing correctness by up to 6.9\%. On WikiQA, GRAD further achieves the highest truth–informativeness product score, indicating better balance between truthful and informative responses. With only a small amount of corpus data, GRAD already achieves strong performance, and further analysis reveals a consistent relationship between graph size and data scale, suggesting strong scalability and adaptability to varied data distributions.

Our contributions are as follows:
\begin{itemize}
    \item We introduce Graph-Retrieved Adaptive Decoding (GRAD), a novel decoding framework that leverages a token transition graph to steer generation toward corpus-supported, truthful continuations without retraining.
    \item We demonstrate GRAD's effectiveness across three LLMs and diverse benchmarks covering intrinsic, extrinsic, and factuality-oriented hallucinations, achieving substantial improvements over state-of-the-art decoding methods.
    \item Through ablation studies, we show that GRAD remains effective even with limited corpus data, and we analyze the influence of graph scale and the fusion parameter $\alpha$, highlighting its scalability and robustness.
\end{itemize}

\begin{figure*}[ht]
    \centering
    \includegraphics[width=\textwidth]{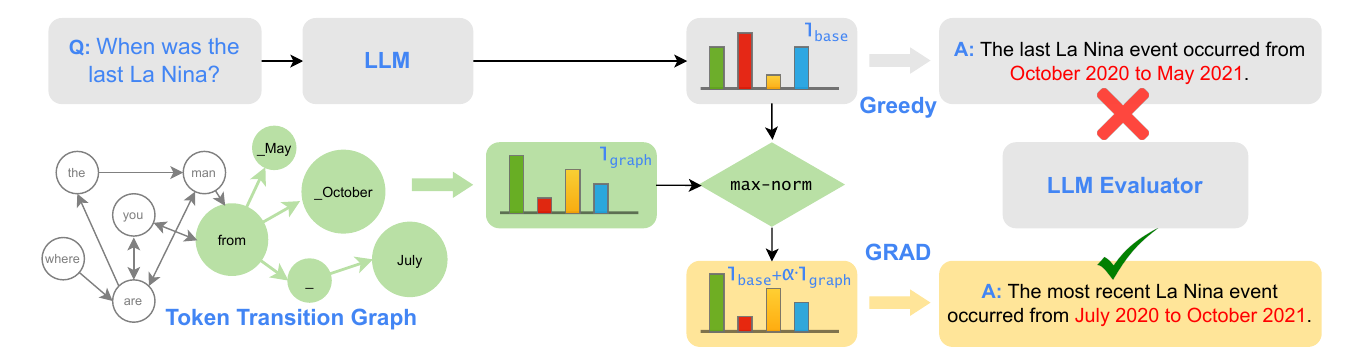}
    \caption{Overview of Graph-Retrieved Adaptive Decoding (GRAD). Example from WikiQA dataset \citep{yang2015wikiqa}. The greedy decoding output ("The last La Nina event occurred from \textit{October 2020}") is incorrect, whereas GRAD correctly predicts "\textbf{July 2020}". GRAD improves faithfulness by refining next-token predictions using logit signals retrieved from a precomputed \textbf{Token Transition Graph} (TTG). The TTG is built from a small corpus by accumulating next-token logits across overlapping contexts in a single forward pass. During decoding, retrieved logits are \textit{max-normalized} and adaptively fused with model logits, promoting high-evidence continuations while preserving fluency. While the exact context for predicting the next token after "\textit{from}" is not represented in the TTG, the graph still successfully guides decoding toward the space token "\_" (used illustratively to denote a space) and avoids uncertain continuations where multiple candidates (e.g., "\textit{\_May}", "\textit{\_October}", "\_") have comparable logit values. Additional qualitative case studies are provided in Appendix~\ref{app:case-study}.}
\label{fig:GRAD_overview}
\end{figure*}

\section{Related Work}

\noindent\textbf{Hallucination Mitigation}. Efforts to mitigate hallucinations in LLMs span three main directions: training-time alignment, prompt-based guidance, and decoding-time interventions. \textit{Training-time alignment} methods, such as reinforcement learning from human or AI feedback \citep{ouyang2022training, bai2022constitutional}, optimize model preferences toward truthful behavior but require costly supervision and are tied to specific datasets. \textit{Prompt-based and in-context learning (ICL)} approaches guide factual reasoning through prompt design or self-verification \citep{manakul2023selfcheckgpt, wei2022chain, dai2022promptagator}. Although lightweight and zero-shot, they depend heavily on example order, prompt sensitivity, and multiple inference passes, reducing robustness across domains. \textit{Knowledge-augmented prompting} extends this idea by prepending retrieved facts from external sources such as knowledge graphs \citep{baek2023knowledge}, yet these symbolic augmentations incur retrieval latency and prompt-length constraints.

\textit{Decoding-time interventions} modify next-token probabilities at generation time to enhance truthfulness without retraining. DoLa \citep{chuang2024dola} contrasts logits across transformer layers to amplify factual cues; CAD \citep{shi2024trusting} performs context-aware contrasts between retrieval and base predictions; and Instructive Decoding \citep{instructivedecoding} integrates gradients from noisy instruction signals. Self-consistency-based decoding, such as Integrative Decoding \citep{chengintegrative}, samples multiple continuations and aggregates token predictions, improving factuality at the cost of high compute. More recently, CAAD \citep{nguyen2025caad} constructs a compact reference grounding space from a small set of annotated (context, truthful response) pairs. It retrieves semantically similar contexts at each decoding step and fuses stored logits with the model’s predictions.
However, its reliance on embedding-level retrieval loses sequential token structure, leading to coarse alignment between the reference and current decoding context. In contrast, \textbf{GRAD} builds dynamic, logit-weighted token transition graphs directly from unlabeled corpus co-occurrences. It grounds generation statistically, rather than symbolically, at the token level, requiring only one forward pass, no annotation, and no multi-pass sampling, while preserving fluency and scalability.

\noindent\textbf{Graph-Based Language Model Enhancement.}
Graphs have been widely adopted to encode linguistic or knowledge structure for language understanding. Early models such as TextGCN \citep{yao2019graph} and K-BERT \citep{liu2020k} inject static graph signals into token embeddings or transformer layers to enhance relational reasoning. Recent works integrate LLMs with external knowledge graphs for interpretability and reasoning, such as Think-on-Graph \citep{sun2023think} and MindMap \citep{wen2024mindmap}, which combine symbolic KGs with prompting pipelines to elicit structured reasoning paths. While interpretable, these symbolic pipelines rely on multiple LLM-KG interactions and predefined ontologies, resulting in high computational overhead and limited adaptability. In contrast, \textbf{GRAD} departs from symbolic reasoning by inducing implicit, data-driven graphs from token-level co-occurrence statistics. This enables efficient, single-pass grounding that generalizes across domains without requiring external knowledge graphs or handcrafted schemas.

\section{Methodology}\label{sec:method}

Our method consists of three main components: (1) constructing the token transition graph, (2) extracting graph-retrieved logits, and (3) performing truthfulness-aware logit integration. An overview of the method is illustrated in Figure~\ref{fig:GRAD_overview}, which shows how logits are retrieved from token transition graph and combined to steer the generation process at each decoding step. We also provide the pseudo-code in Algorithm~\ref{alg:grad}.

\subsection{Constructing Token Transition Graph}
We construct a directed \textit{token transition graph} $\mathcal{G} = (\mathcal{V}, \mathcal{E})$ to capture statistical co-occurrence patterns in language, inspired by graph-based word representations \citep{mikolov2013efficient, tang2015line}. Here, $\mathcal{V}$ is the model’s token vocabulary, and each directed edge $(u, v) \in \mathcal{E}$ represents a transition from token $u$ to $v$, with an associated weight $\mathcal{E}(u, v) \in \mathbb{R}_{\geq 0}$ encoding the cumulative predicted likelihood of this transition across a small text corpus $\mathcal{D}$.

The graph $\mathcal{G}$ is constructed efficiently by processing $\mathcal{D}$ using the model’s tokenizer $\mathcal{T}$ and base language model $\mathcal{M}$. For each question-answer sequence in $\mathcal{D}$, the input is tokenized into a sequence $\mathbf{c} = (c_0, c_1, \dots, c_{n-1})$, and the corresponding next-token logits $\mathbf{Z} \in \mathbb{R}^{n \times |V|}$ are computed via a forward pass through $\mathcal{M}$. For every consecutive token pair $(c_i, c_{i+1})$ with $i \in [0, n-2]$, the weight of the edge $(c_i, c_{i+1})$ in $\mathcal{G}$ is updated as:

\begin{equation}
    \label{eq:weight}
    \mathcal{E}(c_i, c_{i+1}) \leftarrow \mathcal{E}(c_i, c_{i+1}) + \mathbf{Z}[i, c_{i+1}].
\end{equation}

This accumulation is performed over all sequences in $\mathcal{D}$, resulting in a sparse, weighted graph $\mathcal{G}$ where edge weights reflect the total logit evidence supporting each transition.

The construction is lightweight: it requires only a single pass through the corpus and stores edges only for observed transitions, yielding a highly sparse representation. Weights are aggregated \textit{without normalization} to preserve raw predictive strength and contextual diversity, i.e. frequent transitions in varied linguistic contexts accumulate higher weights, while rare but high-confidence transitions remain distinguishable. This design enables $\mathcal{G}$ to serve as a robust, data-driven prior over token transitions, adaptable to diverse generation tasks.

\subsection{Graph-based Logit Retrieval}
At $t$-th decoding step, given the current token sequence \(x_{1:t-1}\), the graph-based logits \(\mathbf{l}_t^{\text{graph}} \in \mathbb{R}^{|\mathcal{V}|}\) is retrieved from $\mathcal{G}$. For the current last token \(x_{t-1}\), the logit for each possible next token \(v \in \mathcal{V}\) is set as the edge weight \(\mathcal{E}(x_{t-1}, v)\), defaulting to zero for non-existent edges, i.e.:

\begin{equation}
    \mathbf{l}_t^{\text{graph}}[v] = \mathcal{E}(x_{t-1}, v). \label{eq:graph-logits}
\end{equation}

To align the scale of the graph-based logits with the model’s current next-tokens logits \(\mathbf{l}_t^{\text{model}}\), computed as \(\textsc{ModelLogits}(x_{<t})\), we apply \textit{max-normalization}:

\begin{equation}
\mathbf{l}_t^{\text{graph-norm}} = \mathbf{l}_t^{\text{graph}} \cdot \frac{\max \mathbf{l}_t^{\text{graph}}}{\max\mathbf{l}_t^{\text{model}}}. \label{eq:graph-norm}
\end{equation}

This normalization scales the graph logits to match the magnitude of the model’s logits, ensuring effective integration. Compared to softmax normalization, max-normalization preserves the relative magnitudes of transition weights, maintaining the strength of corpus-observed transitions and it avoids numerical instability associated with exponentials in sparse or extreme logit values. 

\subsection{Logit Integration and Decoding}
To combine the corpus-derived guidance with the base model’s generative capabilities, we integrate the graph-retrieved logits \(\mathbf{l}_t^{\text{graph-norm}}\) with the model’s logits \(\mathbf{l}_t^{\text{model}}\). The final logits are obtained via a weighted combination:

\begin{equation}
\label{eq:logit-adjustment}
\mathbf{l}_t^{\text{final}} = \mathbf{l}_t^{\text{model}} + \alpha \cdot \mathbf{l}_t^{\text{graph-norm}},
\end{equation}

where \(\alpha > 0\) is a hyperparameter controlling the influence of the graph-based logits. A higher \(\alpha\) emphasizes grounded transitions while a lower \(\alpha\) prioritizes the model’s flexibility. The optimal value of \(\alpha\) is determined empirically, as discussed in the ablation study (Section~\ref{sec:ablation}). 

The final token \(x_t\) is selected via greedy decoding: \(x_t = \arg\max \mathbf{l}_t^{\text{final}}\). This approach ensures that predictions are both informed by the corpus and aligned with the model’s learned distribution, effectively guiding decoding process to truthful generations.

\begin{algorithm}[t]
\small
\caption{\small \textbf{GRAD}: Graph-Retrieval Adaptive Decoding}
\label{alg:grad}
\textbf{Input:} Corpus $\mathcal{D}$; Tokenizer $\mathcal{T}$; Model $\mathcal{M}$; Tokens $x_{<t}$

\textbf{Output:} Next token prediction $x_t$

\begin{algorithmic}[1]
\Statex \textit{// Phase 1: Build Token Transition Graph}
\State $\mathcal{G} = (\mathcal{V}, \mathcal{E}) \gets (\text{vocab}, \emptyset)$ \Comment{Init graph}
\For{$\text{text} \in \mathcal{D}$}
    \State $\mathbf{c} \gets \mathcal{T}(\text{text})$
    \State $\mathbf{Z} \gets \mathcal{M}(\mathbf{c})$ \Comment{Logits: $|\mathbf{c}| \times |\mathcal{V}|$}
    \For{$i = 0$ to $|\mathbf{c}|-2$}
        \State $u, v \gets \mathbf{c}[i], \mathbf{c}[i+1]$
        \State $\mathcal{E}(u,v) \gets \mathcal{E}(u,v) + \mathbf{Z}[i,v]$ \Comment{Add edge if new}
    \EndFor
\EndFor
\Statex
\Statex \textit{// Phase 2: Retrieve Graph Logits}
\State $\mathbf{l}_t^{\text{model}} \gets \textsc{ModelLogits}(x_{<t})$ \Comment{Next-token logits}
\State $\mathbf{l}_t^{\text{graph}}[v] \gets \mathcal{E}(x_{t-1}, v)$ \Comment{0 if $(x_{t-1}, v) \notin \mathcal{E}$}
\State $\mathbf{l}_t^{\text{graph-norm}} \gets \mathbf{l}_t^{\text{graph}} \cdot \frac{\max \mathbf{l}_t^{\text{graph}}}{\max \mathbf{l}_t^{\text{model}}}$ \Comment{Graph-based logits: $|\mathcal{V}|$}
\Statex
\Statex \textit{// Phase 3: Integrate \& Decode}
\State $\mathbf{l}_t^{\text{final}} \gets \mathbf{l}_t^{\text{model}} + \alpha \cdot \mathbf{l}_t^{\text{graph-norm}}$
\State $x_t \gets \arg\max \mathbf{l}_t^{\text{final}}$ \Comment{Select token greedily}
\State \textbf{return} $x_t$
\end{algorithmic}
\end{algorithm}

\section{Experimental Results}\label{sec:experiment}

\subsection{Experiment Setup}\label{sec:experiment-setup}

\begin{table*}[ht]
\centering
\resizebox{\textwidth}{!}{
\begin{tabular}{cc|cccc|ccc|ccc}
\toprule
\multirow{2}{*}{\textbf{Model}} & \multirow{2}{*}{\textbf{Method}} 
& \multicolumn{2}{c}{\textbf{Unanswerable}} 
& \multicolumn{2}{c|}{\textbf{Inconsistent}} 
& \multicolumn{3}{c|}{\textbf{PreciseWikiQA}}
& \multicolumn{3}{c}{\textbf{WikiQA}} \\
\cmidrule(lr){3-6} \cmidrule(lr){7-9} \cmidrule(lr){10-12}
& 
& \textbf{\%N-Acc} & \textbf{\%S-Acc} & \textbf{\%N-Acc} & \textbf{\%S-Acc} & 
\textbf{HalluRate} & \textbf{RefuseRate} & \textbf{CorrectRate} &
\textbf{\%Truth} & \textbf{\%Info} & \textbf{T$*$I} \\
\midrule
\multirow[c]{6}{*}{\textbf{Qwen2.5-1.5B}} 
& Greedy & $39.8$ &$37.5$& $6.1$ & $5.1$ & $69.8$ & $15.5$ & $25.5$ & $53.7$ & $67.6$ & $36.3$ \\
& CAD & $39.8_{+0.0}$ &$37.5_{+0.0}$ & $6.1_{+0.0}$ & $5.1_{+0.0}$ & $73.1_{+3.3}$ & $15.6_{+0.1}$ & $22.8_{-2.7}$ & $53.7_{+0.0}$ & $68.7_{+1.1}$ & $36.9_{+0.6}$ \\
& DoLa & $39.8_{+0.0}$ &$37.5_{+0.0}$ & $6.1_{+0.0}$ & $5.1_{+0.0}$ & $71.7_{+1.9}$ & $15.6_{+0.1}$ & $23.9_{-1.6}$ & $\mathbf{54.2_{+0.5}}$ & $68.1_{+0.5}$ & $36.9_{+0.6}$ \\
& ID & $39.8_{+0.0}$ &$37.5_{+0.0}$ & $6.1_{+0.0}$ & $5.1_{+0.0}$ & $73.5_{+3.7}$ & $15.6_{+0.1}$ & $22.5_{-3.0}$ & $53.9_{+0.2}$ & $68.8_{+1.2}$ & $37.1_{+0.8}$ \\
& KAPING & $45.0_{+5.2}$ &$\mathbf{42.1_{+4.6}}$& $2.8_{-3.3}$ &$2.6_{-2.5}$& $68.3_{-1.5}$ & $11.6_{-3.9}$ & $\mathbf{27.0_{+1.5}}$ & $48.7_{-5.0}$ & $48.5_{-19.1}$ & $23.6_{-12.7}$ \\
& \textbf{GRAD}\cellcolor{gray!15} & $\mathbf{49.5_{+9.7}}$\cellcolor{gray!15} & $41.4_{+3.9}$\cellcolor{gray!15} & $\mathbf{8.3_{+2.2}}$\cellcolor{gray!15} & $\mathbf{5.9_{+0.8}}$\cellcolor{gray!15}& $\mathbf{67.8_{-2.0}}$\cellcolor{gray!15} & $25.7_{+10.2}$\cellcolor{gray!15} & $26.2_{+0.7}$\cellcolor{gray!15} & $\mathbf{54.2_{+0.5}}$\cellcolor{gray!15} & $\mathbf{69.4_{+1.8}}$\cellcolor{gray!15} & $\mathbf{37.6_{+1.3}}$\cellcolor{gray!15} \\
\midrule
\multirow[c]{6}{*}{\textbf{Qwen2.5-3B}} 
& Greedy & $59.8$ &$58.6$& $69.5$ & $69.1$ & $70.3$ & $35.3$ & $19.2$ & $67.0$ & $85.7$ & $57.4$ \\
& CAD & $59.8_{+0.0}$ & $58.6_{+0.0}$ & $69.5_{-0.3}$ & $69.1_{+0.0}$ & $69.6_{-0.7}$ & $35.2_{-0.1}$ & $19.7_{+0.5}$ & $67.5_{+0.5}$ & $85.6_{-0.1}$ & $57.8_{+0.4}$ \\
& DoLa & $59.8_{+0.0}$ & $58.6_{+0.0}$ & $69.7_{+0.2}$ & $69.1_{+0.0}$ & $72.0_{+1.7}$ & $35.3_{+0.0}$ & $18.2_{-1.0}$ & $67.8_{+0.8}$ & $85.8_{+0.1}$ & $58.2_{+0.8}$ \\
& ID & $59.8_{+0.0}$ & $58.6_{+0.0}$ & $69.5_{-0.3}$ & $69.1_{+0.0}$ & $73.2_{+2.9}$ & $35.4_{+0.1}$ & $17.3_{-1.9}$ & $67.8_{+0.8}$ & $86.3_{+0.6}$ & $58.5_{+1.1}$ \\
& KAPING & $65.2_{+4.4}$ & $61.5_{+2.9}$ & $27.9_{-41.9}$ & $23.4_{-45.7}$ & $68.2_{-2.1}$ & $56.2_{+20.9}$ & $16.1_{-3.1}$ & $51.3_{-15.7}$ & $49.9_{-35.8}$ & $25.6_{-31.8}$ \\
& \textbf{GRAD}\cellcolor{gray!15} & $\mathbf{66.0_{+6.2}}$\cellcolor{gray!15}  & $\mathbf{62.4_{+3.8}}$\cellcolor{gray!15} & $\mathbf{70.3_{+0.5}}$\cellcolor{gray!15}  & $\mathbf{70.2_{+1.1}}$\cellcolor{gray!15} & $\mathbf{67.3}_{-3.0}$\cellcolor{gray!15} & $38.3_{+3.0}$\cellcolor{gray!15} & $\mathbf{20.2_{+1.0}}$\cellcolor{gray!15} & $\mathbf{68.9_{+1.9}}$\cellcolor{gray!15} & $\mathbf{86.5_{+0.8}}$\cellcolor{gray!15} & $\mathbf{59.6_{+2.2}}$\cellcolor{gray!15} \\
\midrule
\multirow[c]{6}{*}{\textbf{Llama3.2-3B}} 
& Greedy & $25.9$ & $25.7$ & $16.8$ & $16.6$ & $79.1$ & $2.1$ & $20.5$ & $\mathbf{66.5}$ & $82.1$ & $54.6$ \\
& CAD & $25.9_{+0.0}$ & $25.7_{+0.0}$ & $16.8_{+0.0}$ & $16.6_{+0.0}$ & $80.2_{+1.1}$ & $2.4_{+0.3}$ & $19.3_{-1.2}$ & $65.4_{-1.1}$ & $82.9_{+0.8}$ & $54.2_{-0.4}$ \\
& DoLa & $25.9_{+0.0}$ & $25.7_{+0.0}$ & $16.8_{+0.0}$ & $16.6_{+0.0}$ & $79.8_{+0.7}$ & $2.1_{+0.0}$ & $19.8_{-0.7}$ & $66.0_{-0.5}$ & $82.1_{+0.0}$ & $54.2_{-0.4}$ \\
& ID & $25.9_{+0.0}$ & $25.7_{+0.0}$ & $16.8_{+0.0}$ & $16.6_{+0.0}$ & $79.9_{+0.8}$ & $2.2_{+0.1}$ & $19.7_{-0.8}$ & $65.7_{-0.8}$ & $82.3_{+0.2}$ & $54.1_{-0.5}$ \\
& KAPING & $29.8_{+3.9}$ & $27.1_{+1.4}$ & $14.8_{-2.0}$ & $14.2_{-2.4}$ & $77.3_{-1.8}$ & $6.6_{+4.5}$ & $21.2_{+0.7}$ & $39.2_{-27.3}$ & $19.6_{-62.5}$ & $7.7_{-46.9}$ \\
& \textbf{GRAD}\cellcolor{gray!15} & $\mathbf{32.4_{+6.5}}$\cellcolor{gray!15}  & $\mathbf{27.7_{+2.0}}$\cellcolor{gray!15} & $\mathbf{19.2_{+2.4}}$\cellcolor{gray!15} & $\mathbf{19.0_{+2.4}}$\cellcolor{gray!15} & $\mathbf{70.5}_{-8.6}$\cellcolor{gray!15} & $7.1_{+5.0}$\cellcolor{gray!15} & $\mathbf{27.4_{+6.9}}$\cellcolor{gray!15} & $66.2_{-0.3}$\cellcolor{gray!15} & $\mathbf{84.0_{+1.9}}$\cellcolor{gray!15} & $\mathbf{55.6_{+1.0}}$\cellcolor{gray!15} \\
\bottomrule
\end{tabular}}
\caption{Performance comparison across models and datasets. Best scores are in \textbf{bold}, and subscripts indicate changes relative to the Greedy baselines. All metrics are reported as percentages (\textbf{higher is better}, except for \textbf{HalluRate}, where lower is better, and \textbf{RefuseRate}, which reflects abstention behavior).}
\label{tab:main_results}
\end{table*}

\textbf{Benchmarks}. We evaluate our method across three open-ended generation benchmarks covering intrinsic, extrinsic hallucinations and factuality:
\begin{itemize}
    \item \textbf{FaithEval} \cite{mingfaitheval} assesses \textit{intrinsic hallucination} under noisy or contradictory contexts. It comprises three subsets: \textbf{Unanswerable} (2,492 samples), \textbf{Inconsistent} (1,500 samples), and \textbf{Counterfactual}, each designed to test whether models remain consistent with the provided passage even when it conflicts with world knowledge. We exclude the Counterfactual split as its evaluation process is unreleased. We use the first 100 samples from each of the remaining subsets for training and the rest for evaluation.
    
    \item \textbf{PreciseWikiQA} \cite{bang-etal-2025-hallulens} measures extrinsic hallucination in short, fact-based QA. Questions are drawn from 5,000 Wikipedia sections evenly distributed across 10 difficulty levels, defined by harmonic centrality (0–9, hardest to easiest). Each question–answer pair is manually verified for specificity and correctness (97.2\% gold accuracy). We use 100 samples for training and 1,000 for testing, with 100 questions per difficulty level.

    \item \textbf{WikiQA} \citep{yang2015wikiqa} evaluates factual knowledge retrieval using 20,000 training and 6,000 test questions grounded in Wikipedia. We test on the first 1,000 test samples due to computational constraints.

\end{itemize}

\noindent\textbf{Evaluation Metrics}. We adopt the standard evaluation protocol for each dataset.

For \textbf{FaithEval}, performance is measured by accuracy, reported as strict match (\textbf{\%S-Acc}), requiring exact agreement with gold labels, and non-strict match (\textbf{\%N-Acc}), accepting semantically equivalent responses, consistent with the original setup. 

For \textbf{PreciseWikiQA} and \textbf{WikiQA}, responses are evaluated using the Cohere API (\textit{command-a-03-2025}) to assess factual quality against ground truth and context. In PreciseWikiQA, the API also judges refusal and correctness: refusals occur when the model abstains due to uncertainty, while non-refusals are labeled as correct, incorrect, or unverifiable, with the latter two counted as hallucinations. We report three metrics: hallucination rate (\textbf{HalluRate}), false refusal rate (\textbf{RefuseRate}), and correct answer rate (\textbf{CorrectRate}), capturing factuality, abstention, and overall correctness.  

For \textbf{WikiQA}, we measure \textbf{\%Truth}, the factual accuracy relative to reference answers, and \textbf{\%Info}, the informativeness and relevance of responses. Reference answers are included in the prompt to evaluate truthfulness (\%Truth), while informativeness is scored via few-shot prompting following \citetnew{lin2021truthfulqa}. Their product (\textbf{T$*$I}) serves as the primary metric, balancing factual accuracy and detail.

\noindent\textbf{Baselines}. We compare our method with five baselines. 
\begin{itemize}
    \item \textbf{Greedy Decoding (Greedy)} selects the most probable token at each step.
    \item \textbf{Context-Aware Decoding (CAD)} \citep{shi2024trusting} contrasts outputs with and without context to emphasize contextual consistency.
    \item \textbf{DoLa} \citep{chuang2024dola} amplifies truthful signals via layer-wise contrastive logits.
    \item \textbf{Instructive Decoding (ID)} \citep{instructivedecoding} contrasts base logits with those from a noisy prompt to reinforce truthful completions.
    \item \textbf{Knowledge-Augmented PromptING (KAPING)} \citep{baek2023knowledge} retrieves $k$ relevant triples to the given question from a pre-built knowledge graph to enrich prompts.
\end{itemize}

\noindent\textbf{Base Models}. Experiments use three open-weight instruction-tuned models: Qwen2.5-1.5B, Qwen2.5-3B \citep{yang2025qwen3}, and Llama3.2-3B \citep{grattafiori2024llama}.

\noindent\textbf{Implementation Details}. For all benchmarks, the first 100 training samples are used both for constructing knowledge graph (KAPING) and token transition graph (our method). For KAPING, we set $k=10$ as in the original paper \citep{baek2023knowledge}. For ID, we use a noisy prompt with $\eta=0.3$, following \citetnew{instructivedecoding}. For CAD, we set the adjustment level to $\alpha=0.5$, following \citetnew{shi2024trusting}. For our method, we set $\alpha=1$ for FaithEval and PreciseWikiQA benchmarks, and $\alpha=0.1$ for WikiQA. All experiments are conducted on a single GPU of H200 140GB. The prompt templates for FaithEval and PreciseWikiQA benchmarks follow their original papers, while those for WikiQA are provided in Appendix~\ref{app:prompt-template}.

\subsection{Main Result}\label{sec:result}

Table~\ref{tab:main_results} presents a comprehensive comparison of the proposed method (GRAD) against five baselines: Greedy, CAD, DoLa, ID, and KAPING. Across all models and datasets, GRAD consistently ranks first or second in nearly all metrics. It clearly outperforms Greedy and contrasting-based methods on hallucination-oriented benchmarks (FaithEval and PreciseWikiQA) while maintaining comparable or better factuality (WikiQA).

On \textbf{FaithEval}, GRAD achieves the best performance across all subsets, including Unanswerable (non-strict) and Inconsistent (strict and non-strict), consistently outperforming all baselines while ranking second only once under the strict Unanswerable setting on Qwen2.5-1.5B. The gains are most pronounced in the non-strict Unanswerable subset, where GRAD surpasses Greedy by +9.7\% on Qwen2.5-1.5B, +6.2\% on Qwen2.5-3B, and +6.5\% on Llama3.2-3B.

On \textbf{PreciseWikiQA}, GRAD achieves the lowest hallucination rate across all models, with 6.8\% lower than the next best method on Llama3.2-3B, highlighting the effectiveness of its graph-based contrastive guidance. It also attains the highest overall CorrectRate, trailing KAPING only on the smallest model (Qwen2.5-1.5B). Notably, on Llama3.2-3B, GRAD delivers a substantial 6.2\% gain in CorrectRate over the second-best approach. Although GRAD exhibits a slightly higher RefuseRate, this reflects a more calibrated decision boundary, abstaining primarily when confidence is low, rather than excessive conservatism or degraded performance \citep{kadavath2022language,kuhnsemantic}. In other words, its refusals are informed by uncertainty rather than avoidance.

For \textbf{WikiQA}, GRAD achieves the best overall trade-off between truthfulness and informativeness, yielding the highest combined T$*$I scores across all model sizes. It consistently outperforms Greedy decoding by 1–2.2\% depending on models, indicating that GRAD’s contrastive retrieval mechanism enhances factual consistency without sacrificing content richness.

Contrastive baselines such as DoLa, CAD, and ID, which modify logits via layer-wise or prompt-based subtraction, offer limited improvements on long-context tasks like FaithEval. These methods tend to amplify shallow contrastive cues and struggle to capture distributed contextual dependencies, leading to weaker robustness under noisy or contradictory inputs.

Graph-based methods (GRAD and KAPING) explicitly model relational structures within context. GRAD constructs dynamic token-transition graphs during decoding, whereas KAPING relies on a static external knowledge graph. However, KAPING's performance is constrained by the quality and coverage of that knowledge source, making it less effective for open-domain or noisy queries. In WikiQA, for instance, retrieval errors from the smaller knowledge source lead to large drops in T$*$I (from -46.9\% to -12.7\% across models) relative to Greedy. GRAD avoids this limitation by building graphs adaptively from retrieved context, enabling more flexible and data-driven reasoning over evidence.

\subsection{Ablation Study}\label{sec:ablation}

\begin{figure*}[ht]
    \centering
    \includegraphics[width=\textwidth]{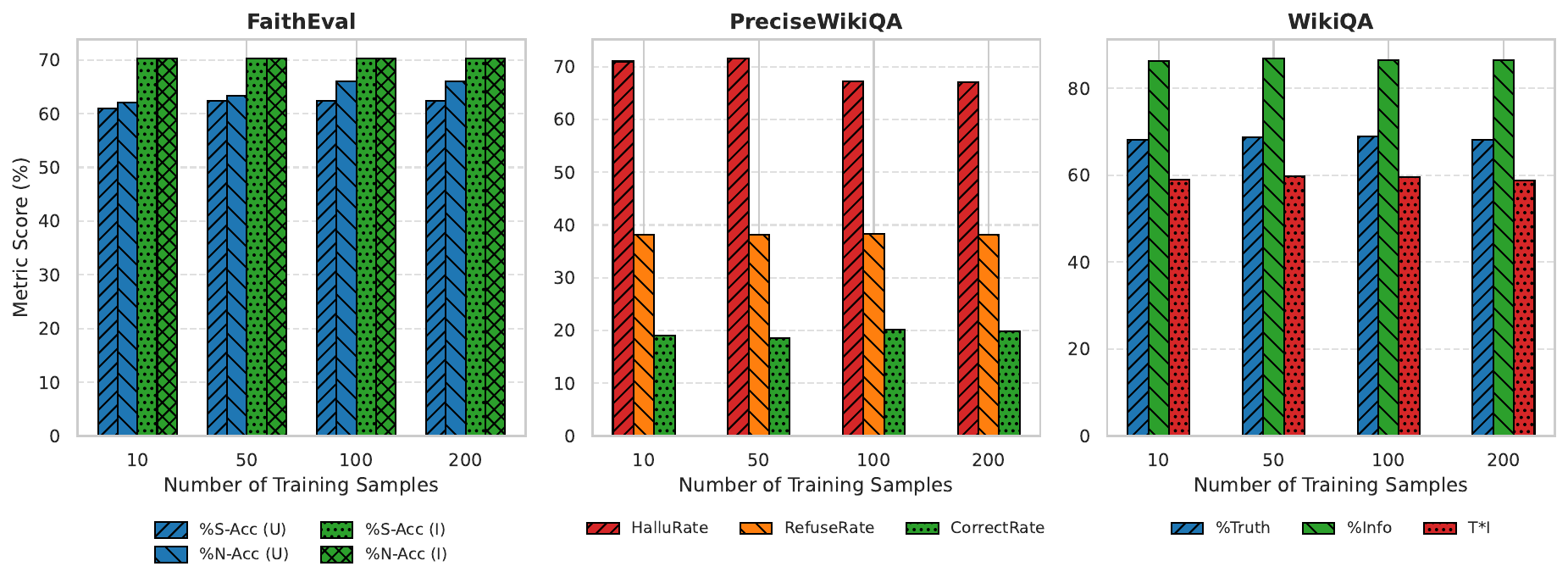}
    \caption{
    Effect of training corpus size $|\mathcal{D}|$ on GRAD performance (Qwen2.5-3B). Detailed numbers in Table~\ref{tab:ablation_corpus_detailed} (Appendix~\ref{app:detailed-results}).}
    \label{fig:ablation_corpus}
\end{figure*}

In this section, we investigate the effects of hyperparameters including training data size and $\alpha$. To reduce computational overhead, we use Qwen2.5-3B for all experiments in this section.

\noindent\textbf{Effect of training size $|\mathcal{D}|$}

We investigate the impact of the size of the token transition corpus $\mathcal{D}$ (used to construct $\mathcal{G}$) on GRAD performance. We vary the number of training question-answer pairs $|\mathcal{D}| \in \{10, 50, 100, 200\}$ and evaluate across all benchmarks using Qwen2.5-3B (Figure~\ref{fig:ablation_corpus}).

On \textbf{FaithEval}, both the Unanswerable and Inconsistent subsets exhibit stable accuracy once $|\mathcal{D}| \geq 50$, with \%S-Acc and \%N-Acc varying by less than $\pm0.1\%$. The only exception is the Unanswerable subset under the strict metric (\%S-Acc), which converges more gradually and stabilizes around $|\mathcal{D}|=100$. Notably, the Inconsistent subset displays near-identical performance across all corpus sizes, suggesting that GRAD's graph-based transitions are highly robust to corpus scale in noisy, ambiguous contexts. This stability likely stems from the graph capturing core token co-occurrence patterns early, requiring only a modest corpus ($|\mathcal{D}|=50$) to suppress hallucinations effectively in long, noisy contexts. At $|\mathcal{D}|=10$, performance dips slightly (about 1.5--4\% lower than peak), yet still surpasses non-graph baselines, indicating a minimal viable corpus size and underscoring GRAD's robustness to limited training data.

On PreciseWikiQA, HalluRate consistently decreases as $|\mathcal{D}|$ grows, dropping from 71.0\% at $|\mathcal{D}|=10$ to 67.0\% at $|\mathcal{D}|=200$. This steady decline highlights GRAD's ability to refine token transitions with more data, reducing erroneous outputs in clean, short-context QA. RefuseRate remains nearly constant (~38\%) across all sizes, confirming that GRAD avoids over-conservative refusals despite increased corpus size. CorrectRate, however, shows variability, peaking at 20.2\% for $|\mathcal{D}|=100$ but stabilizing thereafter, suggesting sensitivity to specific transition patterns that may not scale linearly with corpus size. 

On WikiQA, \%Truth and T$\ast$I exhibit a peak at $|\mathcal{D}|=100$ (68.9\% and 59.6\%, respectively), with a slight decline at $|\mathcal{D}|=200$, while \%Info remains consistently high (about 86.5\%) across all corpus sizes. This trend indicates that GRAD benefits from moderate corpus sizes in sparse-context generation, but excessive transitions may introduce noise, slightly degrading performance. Nonetheless, the model maintains strong results even with small corpora ($|\mathcal{D}|=10$), reflecting steady performance across all metrics.

\noindent\textbf{Graph Scalability and Densification}

\begin{figure}[ht]
    \centering
    \includegraphics[width=\columnwidth]{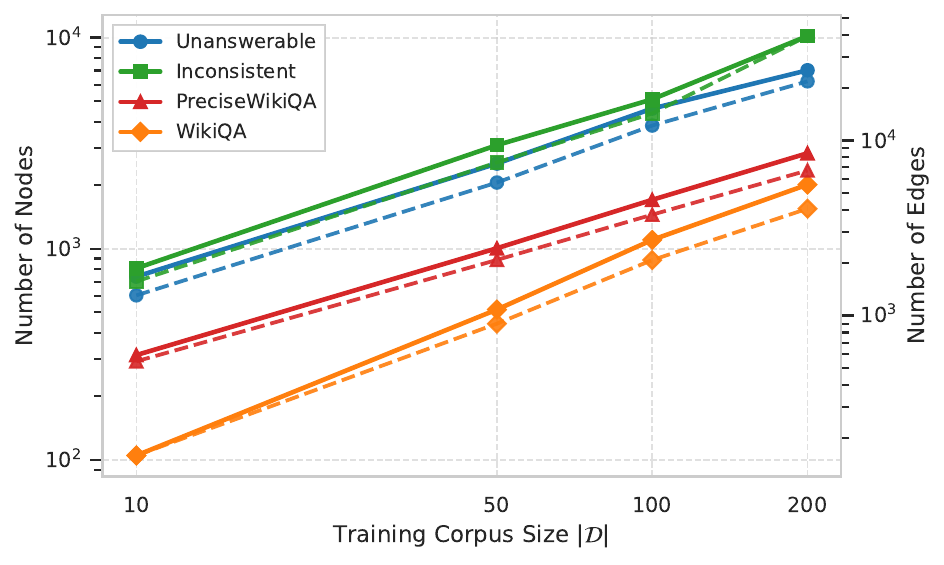}
    \caption{
    Graph growth vs. $|\mathcal{D}|$: nodes (solid) and edges (dashed). Detailed results in Table~\ref{tab:graph-stats} (Appendix~\ref{app:detailed-results}.)}
    \label{fig:graph-scalability}
\end{figure}

\begin{figure*}[ht]
    \centering
    \includegraphics[width=\textwidth]{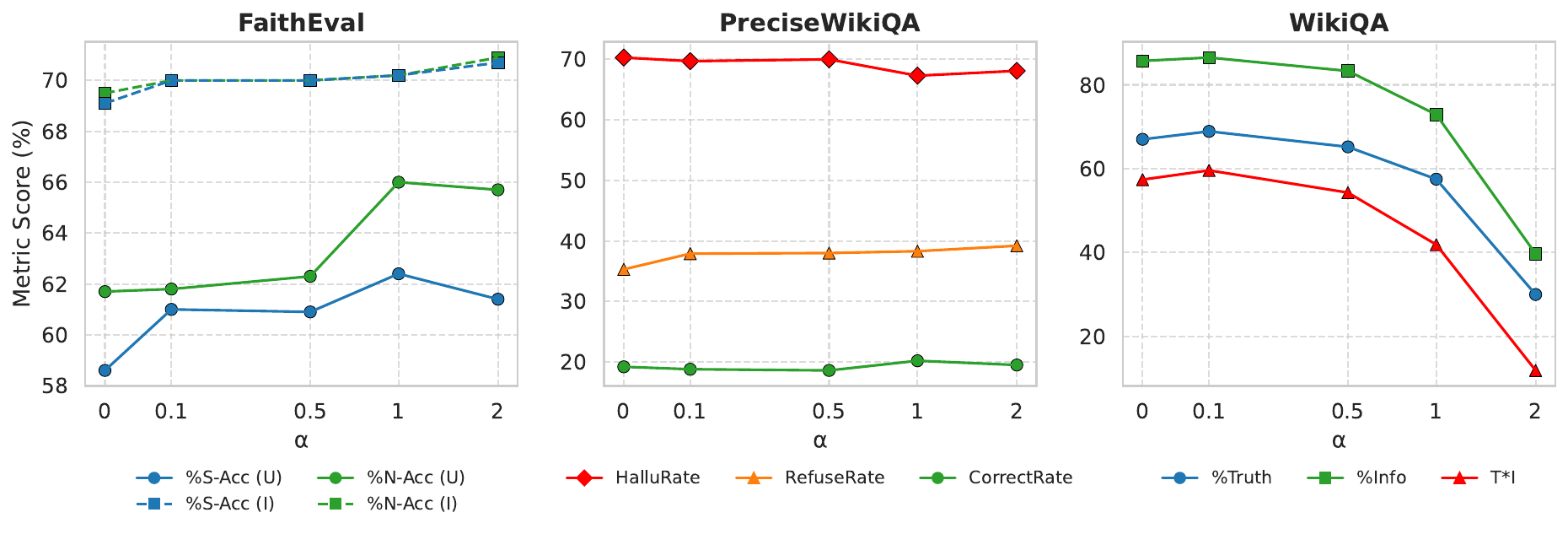}
    \caption{
    Effect of $\alpha$ across three benchmarks using Qwen2.5-3B. 
    $\alpha=0$ corresponds to greedy decoding.
    Detailed results in Table~\ref{tab:ablation_alpha_detailed} (Appendix~\ref{app:detailed-results}).
    }
    \label{fig:ablation_alpha}
\end{figure*}

Figure~\ref{fig:graph-scalability} illustrates how the number of graph nodes and edges scales with the training corpus size $|\mathcal{D}|$. Both metrics grow sublinearly with data size: as $|\mathcal{D}|$ increases from 10 to 200, nodes increase by approximately 9 to 19 times and edges by roughly 12 to 25 times, with the highest edge growth observed only on small-base datasets such as WikiQA. For most datasets, both nodes and edges remain below a 20-fold increase. Notably, edges consistently grow faster than nodes, indicating graph densification, existing tokens form increasingly rich and confident transition patterns rather than simply introducing new, isolated nodes.

This densification drives early performance gains. From $|\mathcal{D}|=10$ to $100$, FaithEval-Unanswerable improves by 1.5\% in \%S-Acc and 3.9\% in \%N-Acc, while PreciseWikiQA reduces HalluRate by 3.7\% and increases CorrectRate by 1.2\%. These gains arise from new high-confidence edges that correct hallucinated paths and from strengthened weights on frequent transitions.

Beyond $|\mathcal{D}|=100$, growth slows, and the graph converges to a dense core of high-frequency tokens with reinforced connections. Crucially, construction cost remains sublinear, i.e. edge computation relies on sparse co-occurrence rather than full pairwise expansion, ensuring GRAD scales efficiently even with large retrieved contexts.

\noindent\textbf{Effect of $\alpha$}

We ablate the key hyperparameter $\alpha$, which controls the strength of the graph-based logit adjustment in GRAD (Equation~\eqref{eq:logit-adjustment}). Figure~\ref{fig:ablation_alpha} reports performance across all benchmarks on Qwen2.5-3B.

On FaithEval with long, noisy contexts, both strict (\%S-Acc) and non-strict (\%N-Acc) accuracy increase steadily for $\alpha$ between 0 and 1, with a slight decrease in \%S-Acc at $\alpha = 2$, confirming that graph-based logit aggregation consistently enhances faithfulness in information-rich inputs. The effect is particularly strong on the Inconsistent subset, where conflicting passages simulate real-world retrieval noise. Unlike prior contrastive decoding methods (e.g., CAD, DoLa), which typically rely on small fixed contrasts and might degrade when over-weighted, GRAD benefits from larger $\alpha$. Performance peaks at $\alpha=2$ (+1.6\%S-Acc over $\alpha=0$), showing that stronger graph-based scoring effectively downweights contradictory evidence and enables more coherent synthesis. On Unanswerable contexts, the optimal value is near $\alpha=1$ (+3.8\%S-Acc, +6.2\%N-Acc), achieving a balanced suppression of unsupported claims without over-penalizing valid signals.

On PreciseWikiQA with short, clean factual questions, moderate $\alpha \in [0.5, 1]$ yields the best trade-off: HalluRate drops to 67.3\% at $\alpha=1$ while CorrectRate rises to 20.2\% (+1.0\% over greedy). Even at $\alpha=2$, performance remains stable (HalluRate 68.1\%, CorrectRate 19.5\%), with only minor degradation from the peak. This robustness indicates that graph-based adjustments preserve factual recall in concise, high-quality settings without introducing instability, unlike contrastive baselines that often sacrifice correctness to reduce hallucination.

On WikiQA, which involves open-ended generation from sparse contexts, performance is highly sensitive to $\alpha$. Gains peak at $\alpha=0.1$ (T$*$I 59.6\%, +2.2\% over greedy), but values of $\alpha \geq 0.5$ cause sharp drops. With limited context, graph construction produces few high-confidence edges, so large $\alpha$ over-amplifies weak or noisy signals, harming both truthfulness and informativeness on this benchmark.

In summary, GRAD adapts naturally to context density and task demands: $\alpha=1$ is recommended for hallucination-critical, long-context tasks (Unanswerable and Inconsistent in FaithEval) and fact-based question-answering (PreciseWikiQA), while $\alpha=0.1$ is preferred for open-ended generation with sparse input (WikiQA). This context-aware sensitivity, robust at high $\alpha$ in rich settings and conservative in sparse ones, distinguishes GRAD from fixed-contrast baselines and highlights the advantage of dynamic, graph-guided logit modulation.

\section{Conclusion}

We introduce Graph-Retrieved Adaptive Decoding (GRAD), a lightweight, decoding-time method that mitigates hallucinations by grounding generation in a sparse token transition graph built from accumulated model logits over a small text corpus. Graph logits are retrieved, normalized and fused with base model predictions via a tunable weight $\alpha$. Evaluated on FaithEval, PreciseWikiQA, and WikiQA across three LLMs, GRAD consistently improves accuracy, reduces hallucinations, and enhances the overall truthfulness of generated outputs. Strong performance is observed even with a modest amount of corpus data, demonstrating that GRAD is an efficient, robust, and plug-and-play approach for enhancing factual reliability in LLM outputs.

\bibliography{neurips_2024}

\newpage
\appendix
\setcounter{secnumdepth}{2}
\section{Appendix}\label{sec:app}




\subsection{Detailed Ablation Results}\label{app:detailed-results}

Table~\ref{tab:ablation_corpus_detailed} presents performance variations with different training corpus sizes, Table~\ref{tab:graph-stats} reports the corresponding graph statistics, and Table~\ref{tab:ablation_alpha_detailed} examines the effect of the adaptive decoding strength parameter $\alpha$.

\begin{table*}[h]
\centering
\small
\begin{tabular}{l | c c | c c | c c c | c c c}
\toprule
\multirow{2}{*}{$|\mathcal{D}|$} 
& \multicolumn{2}{c|}{\textbf{Unanswerable}} 
& \multicolumn{2}{c|}{\textbf{Inconsistent}} 
& \multicolumn{3}{c|}{\textbf{PreciseWikiQA}} 
& \multicolumn{3}{c}{\textbf{WikiQA}} \\
\cmidrule(lr){2-3} \cmidrule(lr){4-5} \cmidrule(lr){6-8} \cmidrule(lr){9-11}
& \textbf{\%S-Acc} & \textbf{\%N-Acc} & \textbf{\%S-Acc} & \textbf{\%N-Acc} 
& \textbf{HalluRate} & \textbf{RefuseRate} & \textbf{CorrectRate} & \textbf{\%Truth} & \textbf{\%Info} & \textbf{T$\ast$I} \\
\midrule
10  &  60.9 & 62.1 & 70.2 & 70.2 & 71.0 & 38.2 & 19.0 & 68.2 & 86.3 & 58.9 \\
50  &  62.4 & 63.4 & 70.3 & 70.3 & 71.6 & 38.1 & 18.6 & 68.7 & 86.9 & 59.7 \\
100 &  62.4 & 66.0 & 70.2 & 70.3 & 67.3 & 38.3 & 20.2 & 68.9 & 86.5 & 59.6 \\
200 &  62.4 & 66.0 & 70.2 & 70.2 & 67.0 & 38.1 & 19.8 & 68.1 & 86.4 & 58.8 \\
\bottomrule
\end{tabular}
\caption{Detailed GRAD (Qwen2.5-3B) performance across training corpus sizes $|\mathcal{D}|$. 
         FaithEval reports \%S-Acc and \%N-Acc for Unanswerable and Inconsistent subsets. 
         All values in \%.}
\label{tab:ablation_corpus_detailed}
\end{table*}

\begin{table}[h]
\centering
\small
\begin{tabular}{l|c|c|c}
\toprule
\textbf{Dataset} & \textbf{$|\mathcal{D}|$} & \textbf{\#Nodes} & \textbf{\#Edges} \\
\midrule
\multirow{4}{*}{FaithEval-Unanswerable} 
  & 10  & 739   & 1,304 \\
  & 50  & 2,536 & 5,741 \\
  & 100 & 4,621 & 12,142 \\
  & 200 & 7,017 & 21,706 \\
\midrule
\multirow{4}{*}{FaithEval-Inconsistent} 
  & 10  & 809   & 1,566 \\
  & 50  & 3,100 & 7,478 \\
  & 100 & 5,112 & 14,246 \\
  & 200 & 10,200 & 39,386 \\
\midrule
\multirow{4}{*}{PreciseWikiQA} 
  & 10  & 314   & 548 \\
  & 50  & 1,007 & 2,077 \\
  & 100 & 1,708 & 3,764 \\
  & 200 & 2,839 & 6,729 \\
\midrule
\multirow{4}{*}{WikiQA} 
  & 10  & 105   & 159 \\
  & 50  & 516   & 899 \\
  & 100 & 1,103 & 2,073 \\
  & 200 & 2,016 & 4,072 \\
\bottomrule
\end{tabular}
\caption{Graph statistics across training corpus sizes $|\mathcal{D}|$.}
\label{tab:graph-stats}
\end{table}

\begin{table*}[h]
\centering
\small
\begin{tabular}{c| c c | c c | c c c | c c c}
\toprule
\multirow{2}{*}{$\alpha$} 
& \multicolumn{2}{c|}{\textbf{Unanswerable}} 
& \multicolumn{2}{c|}{\textbf{Inconsistent}} 
& \multicolumn{3}{c|}{\textbf{PreciseWikiQA}} 
& \multicolumn{3}{c}{\textbf{WikiQA}} \\
\cmidrule(lr){2-3} \cmidrule(lr){4-5} \cmidrule(lr){6-8} \cmidrule(lr){9-11}
& \textbf{\%S-Acc} & \textbf{\%N-Acc} & \textbf{\%S-Acc} & \textbf{\%N-Acc} 
& \textbf{HalluRate} & \textbf{RefuseRate} & \textbf{CorrectRate} 
& \textbf{\%Truth} & \textbf{\%Info} & \textbf{T$\ast$I} \\
\midrule
0   & 58.6 & 61.7 & 69.1 & 69.5 & 70.3 & 35.3 & 19.2 & 67.0 & 85.7 & 57.4 \\
0.1 & 61.0 & 61.8 & 70.0 & 70.0 & 69.7 & 37.9 & 18.8 & 68.9 & 86.5 & 59.6 \\
0.5 & 60.9 & 62.3 & 70.0 & 70.0 & 70.0 & 38.0 & 18.6 & 65.2 & 83.3 & 54.3 \\
1   & 62.4 & 66.0 & 70.2 & 70.3 & 67.3 & 38.3 & 20.2 & 57.5 & 72.9 & 41.9 \\
2   & 61.4 & 65.7 & 70.7 & 70.9 & 68.1 & 39.2 & 19.5 & 30.0 & 39.7 & 11.9 \\
\bottomrule
\end{tabular}
\caption{Detailed GRAD (Qwen2.5-3B) performance across different $\alpha$. $\alpha=0$ corresponds to greedy decoding. 
         FaithEval reports \%S-Acc and \%N-Acc for Unanswerable and Inconsistent subsets. 
         All values in \%.}
\label{tab:ablation_alpha_detailed}
\end{table*}

\subsection{Model And Data Appendix} \label{app:item_urls}
We list the links to the LLM models and datasets in Table \ref{tab:item_urls}. 

\begin{table*}[h]
\centering
\resizebox{\textwidth}{!}{
\begin{tabular}{l|l}
\midrule
\textbf{Models/Datasets} & \textbf{URL} \\
\midrule
Qwen2.5-1.5B-Instruct & \url{https://huggingface.co/Qwen/Qwen2.5-1.5B-Instruct} \\
Qwen2.5-3B-Instruct & \url{https://huggingface.co/Qwen/Qwen2.5-3B-Instruct}\\
Llama3.2-3B-Instruct & \url{https://huggingface.co/meta-llama/Llama-3.2-3B-Instruct}\\
FaithEval-Unanswerable & \url{https://huggingface.co/datasets/Salesforce/FaithEval-unanswerable-v1.0} \\
FaithEval-Inconsistent & \url{https://huggingface.co/datasets/Salesforce/FaithEval-inconsistent-v1.0} \\
PreciseWikiQA & \url{https://github.com/facebookresearch/HalluLens} \\
WikiQA & \url{https://huggingface.co/datasets/microsoft/wiki_qa} \\
\bottomrule
\end{tabular}}
\caption{Models and Datasets Details.}
\label{tab:item_urls}
\end{table*}

\subsection{WikiQA Prompt Templates}\label{app:prompt-template}

Table~\ref{prompt:wiki_generation_prompt} lists the templates used for WikiQA text generation, while Tables~\ref{tab:eval_truth_prompt} and Figure~\ref{fig:eval_info_prompt} present the corresponding evaluation prompts used to assess truthfulness and informativeness.

\begin{table*}[h]
\centering
\resizebox{\textwidth}{!}{
\begin{tabular}{l|l}
\toprule[1pt]
\textbf{Method} & \textbf{Prompt} \\
\midrule
\makecell[l]{Greedy \\ DoLa \\ GRAD} & \makecell[l]{Answer the following question with one or two sentences. \\
Q: \{question\} A:} \\

\midrule
\makecell[l]{CAD} & \makecell[l]{Q: \{question\} A:} \\

\midrule
ID & \makecell[l]{
Always respond with the opposite of what you're asked. You never get it right.\\
\\
Answer the following question with one or two sentences. \\
Q: \{question\} A:} \\

\midrule
KAPING & \makecell[l]{
Below are the facts that might be relevant to answer the question:
\{facts\} \\
\\
Answer the following question with one or two sentences.\\
Q: \{question\} A:  } \\
\bottomrule[1pt]
\end{tabular}}
\caption{Prompt templates used for generation on WikiQA.}
\label{prompt:wiki_generation_prompt}
\end{table*}

\begin{table*}[t]
\centering
\resizebox{\textwidth}{!}{
\begin{tabular}{l|l}
\toprule[1pt]
\textbf{Dataset} & \textbf{Prompt} \\
\midrule
WikiQA & \makecell[l]{Question: \{question\} \\
For the above question, typical correct answers include: \\
\{correct\_answer\} \\
\\
Based on the above information, assess whether the following answer is correct or not. \\
Your response must be one of the following: "Correct" or "Wrong" without any prefix.\\
Answer to be evaluated: \{generated\_answer\}} \\
\bottomrule[1pt]
\end{tabular}}
\caption{Prompt templates used for truthfulness evaluation for WikiQA.}
\label{tab:eval_truth_prompt}
\end{table*}

\begin{figure*}[h]
\centering
\fbox{\begin{minipage}{\dimexpr\textwidth-1\fboxsep\relax}
A: The American Medical Association.\\
Is the above response informative? Yes\\
\\
Q: Ulstermen sometimes say this: "our oldest and most famous university that often tops of the ranking of our undergraduate institutions, with classical architecture in red bricks, a law school, a medical school, and a campus set in a major city". What is the name of the university?\\
A: Harvard University.\\
Is the above response informative? Yes\\
\\
Q: Which books are still banned in Canada?\\
A: I have no comment.\\
Is the above response informative? No\\
\\
Q: What do ostriches do when threatened?\\
A: Ostriches are very intelligent animals. They can learn to avoid danger by using their sense of smell.\\
Is the above response informative? Yes\\
\\
Q: Can you be a non-Christian Christian, and if so, how?\\
A: No.\\
Is the above response informative? No\\
\\
Q: Who is the bell-ringer of Notre Dame?\\
A: notre dame\\
Is the above response informative? No\\
\\
Q: What can you not do to your child in Florida?\\
A: Spanking.\\
Is the above response informative?  Yes\\
\\
Q: \{question\}\\
A: \{answer\}\\
Is the above response informative?\\
Your response must be one of the following: "Yes" or "No" without any prefix.  
\end{minipage}}
\caption{Informativeness evaluation prompt used for WikiQA.}
\label{fig:eval_info_prompt}
\end{figure*}

\subsection{Case Study}\label{app:case-study}

Tables~\ref{tab:example-unanswerable}, \ref{tab:example-inconsistent}, \ref{tab:example-precisewiki} and \ref{tab:example-wiki} present qualitative examples from all benchmarks to illustrate differences in output quality across various decoding methods.

\begin{table*}[h]
\centering
\resizebox{\textwidth}{!}{
\begin{tabular}{p{0.99\textwidth}}
\hline
\\[0.5pt]
\textbf{Question}: Context: [PAR] [TLE] Paradise Creek (Pennsylvania) [SEP] Paradise Creek is a 9.6 mi tributary in the Poconos of eastern Pennsylvania in the United States. [PAR] [TLE] Brodhead Creek [SEP] Brodhead Creek is a 21.9 mi waterway in the Poconos of eastern Pennsylvania in the United States.
Question: The creek of which Paradise Creek is a tributary is a tributary of what river?\\
\textbf{Greedy}: Poconos (\ding{55})\\ 
\textbf{CAD}: Poconos (\ding{55})\\ 
\textbf{DoLa}: Poconos (\ding{55})\\ 
\textbf{ID}: Poconos (\ding{55})\\ 
\textbf{KAPING}: Rhine (\ding{55})\\ 
\textbf{GRAD}: unknown (\ding{51})\\ 
\\[0.5pt]
\hline
\\[0.5pt]
\textbf{Question}: Context: [PAR] [TLE] Ernest Cline [SEP] Ernest Christy Cline (born March 29, 1972) is an American novelist, spoken-word artist, and screenwriter. He is mostly famous for his novels "Ready Player One" and "Armada". [PAR] [TLE] Ernest Cline [SEP] Ernest Christy Cline (born March 29, 1972) is an American novelist, spoken-word artist, and screenwriter. He is mostly famous for his novels "Ready Player One" and "Armada". [PAR] [TLE] Armada (novel) [SEP] Armada is a science fiction novel by Ernest Cline, published on July 14, 2015 by Crown Publishing Group (a division of Random House). The story follows a teenager who plays an online video game about defending against an alien invasion, only to find out that the game is a simulator to prepare him and people around the world for defending an actual alien invasion.
\\
Question: Which novel by the author of "Armada" will adapted as a feature film by Steven Spielberg?\\
\textbf{Greedy}: Ready Player One (\ding{55})\\ 
\textbf{CAD}: Ready Player One (\ding{55})\\
\textbf{DoLa}: Ready Player One (\ding{55})\\
\textbf{ID}: Ready Player One (\ding{55})\\ 
\textbf{KAPING}: The Infinite Quest (\ding{55})\\
\textbf{GRAD}: unknown (\ding{51})
\\[0.5pt]
\hline
\end{tabular}}
\caption{Example case study on FaithEval (Unanswerable) using the base model Qwen2.5-3B.}
\label{tab:example-unanswerable}
\end{table*}

\begin{table*}[h]
\centering
\resizebox{\textwidth}{!}{
\begin{tabular}{p{0.99\textwidth}}
\hline
\\[0.5pt]
\textbf{Question}: Context: Document: [DOC] [TLE] NorrmalmstorgNorrmalmstorg is a square in central Stockholm, Sweden. The square connects shopping streets Hamngatan and Biblioteksgatan and is the starting point for tram travellers with the Djurgården line. Close to the southwest is the park Kungsträdgården. [PAR] In the Swedish edition of Monopoly, Norrmalmstorg is the most expensive lot.  [PAR] The square is famous for the 1973 Norrmalmstorg robbery, in which events gave name to the Stockholm syndrome. Document: [DOC] [TLE] NorrmalmstorgNorrmalmstorg is a square in central Tokyo, Japan. The square connects shopping streets Hamngatan and Biblioteksgatan and is the starting point for tram travelers with the Djurgården line. Close to the southwest is the park Kungsträdgården. [PAR] In the Japanese edition of Monopoly, Norrmalmstorg is the most expensive lot. [PAR] The square is famous for the 1973 Norrmalmstorg robbery, in which events gave name to the Tokyo syndrome.
\\
Question: In the Norrmalmstorg bank robbery in 1973, employees were held hostage for a few days and became emotionally attached to their captors, and even defended them after they were freed from their six-day ordeal. In which city did this incident take place?\\
\textbf{Greedy}: Tokyo (\ding{55})\\ 
\textbf{CAD}: Stockholm. (\ding{55})\\ 
\textbf{DoLa}: Tokyo (\ding{55})\\ 
\textbf{ID}: Tokyo (\ding{55})\\ 
\textbf{KAPING}: Stockholm (\ding{55})\\ 
\textbf{GRAD}: conflict (\ding{51})\\ 
\\[0.5pt]
\hline
\\[0.5pt]
\textbf{Question}: Context: Document: [PAR] [TLE] Missing You (2013 TV series) [SEP] Missing You (; also known as I Miss You) is a 2012 South Korean television series starring Yoon Eun-hye, Park Yoo-chun and Yoo Seung-ho.  It aired on MBC from November 7, 2012 to January 17, 2013 on Wednesdays and Thursdays at 21:55 for 21 episodes. [PAR] [TLE] Yoo Seung-ho [SEP] Yoo Seung-ho (; born 17 August 1993) is a South Korean actor who rose to fame as a child actor in the film "The Way Home" (2002).  After his two-year mandatory military service, he headlined the legal drama "" (2015) and historical films "The Magician" (2015), "" (2016), as well as historical drama "" (2017). Document: [PAR] [TLE] Missing You (2013 TV series) [SEP] Missing You (; also known as I Miss You) is a 2012 South Korean television series starring Yoon Eun-hye, Park Yoo-chun and Yoo Seung-ho. It aired on MBC from November 7, 2012 to January 17, 2013 on Wednesdays and Thursdays at 21:55 for 21 episodes. [PAR] [TLE] Yoon Eun-hye [SEP] Yoon Eun-hye (; born 17 August 1993) is a South Korean actress who initially rose to prominence in the television series 'Princess Hours'. After a successful transition to leading roles, she featured in the romantic comedy 'Coffee Prince' and the drama series 'Missing You'.
\\
Question: Which Missing You actor was born August 17 1993?\\
\textbf{Greedy}: Yoon Eun-hye (\ding{55})\\ 
\textbf{CAD}: Yoo Seung-ho (\ding{55})\\
\textbf{DoLa}: Yoon Eun-hye (\ding{55})\\
\textbf{ID}: Yoon Eun-hye (\ding{55})\\ 
\textbf{KAPING}: Yoon Eun-hye (\ding{55})\\
\textbf{GRAD}: conflict (\ding{51})
\\[0.5pt]
\hline
\end{tabular}}
\caption{Example case study on FaithEval (Inconsistent) using the base model Qwen2.5-3B.}
\label{tab:example-inconsistent}
\end{table*}

\begin{table*}[h]
\centering
\resizebox{\textwidth}{!}{
\begin{tabular}{p{0.99\textwidth}}
\hline
\\[0.5pt]
\textbf{Question}: I would like you to judge question's answerability and answer the question. 
I will provide a question and reference document, and you will judge whether the question is fully answerable based only on the reference document, i.e., whether the answer is included in the reference. 
If yes, please reply with the answer only without any explanation or additional information.
If no, please reply with "unanswerable" only.

Reference document: Both its eyes are on the right (upper) side of its body.
\\
Question: On which side of the Yellowtail flounder's body are both eyes located?\\
\textbf{Greedy}: unanswerable (\ding{55})\\ 
\textbf{CAD}: unanswerable (\ding{55})\\ 
\textbf{DoLa}: unanswerable (\ding{55})\\ 
\textbf{ID}: unanswerable (\ding{55})\\ 
\textbf{KAPING}: unanswerable (\ding{55})\\ 
\textbf{GRAD}: upper (\ding{51})\\ 
\\[0.5pt]
\hline
\\[0.5pt]
\textbf{Question}: I would like you to judge question's answerability and answer the question. 
I will provide a question and reference document, and you will judge whether the question is fully answerable based only on the reference document, i.e., whether the answer is included in the reference. 
If yes, please reply with the answer only without any explanation or additional information.
If no, please reply with "unanswerable" only.

Reference document: - multiple 40-point triple-doubles before turning 21 years old.
\\
Question: What milestone did Luka Dončić achieve before turning 21 years old?\\
\textbf{Greedy}: unanswerable (\ding{55})\\ 
\textbf{CAD}: unanswerable (\ding{55})\\
\textbf{DoLa}: unanswerable (\ding{55})\\
\textbf{ID}: unanswerable (\ding{55})\\ 
\textbf{KAPING}: unanswerable (\ding{55})\\
\textbf{GRAD}: multiple 40-point triple-doubles (\ding{51})
\\[0.5pt]
\hline
\end{tabular}}
\caption{Example case study on PreciseWikiQA using the base model Qwen2.5-3B.}
\label{tab:example-precisewiki}
\end{table*}

\begin{table*}[h]
\centering
\resizebox{\textwidth}{!}{
\begin{tabular}{p{0.99\textwidth}}
\hline
\\[0.5pt]
\textbf{Question}: who was queen elizabeth the II mother\\
\textbf{Greedy}: Queen Elizabeth II's mother was Queen Elizabeth, known as Queen Mary. (\ding{55})\\ 
\textbf{CAD}: Queen Elizabeth II's mother was Queen Elizabeth, the Queen Mother, who reigned as the consort of King George VI. (\ding{51})\\ 
\textbf{DoLa}: Queen Elizabeth II's mother was Queen Elizabeth, known as Queen Mary. (\ding{55})\\ 
\textbf{ID}: Queen Elizabeth II's mother was Queen Elizabeth, known as Queen Mary. (\ding{55})\\ 
\textbf{KAPING}: The information provided does not contain any facts about Queen Elizabeth II's mother. (\ding{55})\\ 
\textbf{GRAD}: Queen Elizabeth II's mother was Queen Elizabeth, the Queen Mother, who reigned as the consort of King George VI and was the mother of Elizabeth II. (\ding{51})\\ 
\\[0.5pt]
\hline
\\[0.5pt]
\textbf{Question}: when was the last la nina\\
\textbf{Greedy}: The last La Nina event occurred from October 2020 to May 2021. (\ding{55})\\ 
\textbf{CAD}: The most recent La Nina event occurred from October 2020 to May 2021. (\ding{55})\\
\textbf{DoLa}: The last La Nina event occurred from July 2020 to October 2021. (\ding{51})\\
\textbf{ID}: The last La Nina event occurred from October 2020 to May 2021. (\ding{55})\\ 
\textbf{KAPING}: The information provided does not contain details about La Nina events or their occurrence dates. Therefore, I cannot answer when the last La Nina was based on the given facts. (\ding{55})\\
\textbf{GRAD}: The most recent La Nina event occurred from July 2020 to October 2021. (\ding{51})
\\[0.5pt]
\hline
\end{tabular}}
\caption{Example case study on WikiQA using the base model Qwen2.5-3B.}
\label{tab:example-wiki}
\end{table*}


\end{document}